\documentclass[10pt,twocolumn,letterpaper]{article}

\usepackage{wacv}
\usepackage{times}
\usepackage{epsfig}
\usepackage{graphicx}
\usepackage{amsmath}
\usepackage{amssymb}
\usepackage{url}
\usepackage{multirow}
\usepackage{enumitem}

\usepackage[pagebackref=true,breaklinks=true,colorlinks,bookmarks=false]{hyperref}

\wacvfinalcopy 

\def\wacvPaperID{926} 

\newcommand{\B}[1]{\textbf{#1}}
\newcommand{\name}{action-sequence-classification}
\newcommand{\Name}{Action-sequence-classification}

\ifwacvfinal\fi
\begin{document}

\title{Human Action Sequence Classification}

\author{Yan Bin Ng\\
A*AI, A*STAR\\ Singapore.\\
{\tt\small \url{ng_yan_bin@scei.a-star.edu.sg}}
\and
Basura Fernando \\
A*AI, A*STAR\\ Singapore.\\
{\tt\small \url{fernando_basura@scei.a-star.edu.sg, basuraf@gmail.com}}
}

\maketitle
\ifwacvfinal\thispagestyle{empty}\fi
\begin{abstract}
	This paper classifies human action sequences from videos using a machine translation model. In contrast to classical human action classification which outputs \emph{a set of actions}, our method output a \emph{sequence of action} in the chronological order of the actions performed by the human. Therefore our method is evaluated using sequential performance measures such as Bilingual Evaluation Understudy (BLEU) scores. Action sequence classification has many applications such as learning from demonstration, action segmentation, detection, localization and video captioning.
Furthermore, we use our model that is trained to output action sequences to solve downstream tasks; such as video captioning and action localization. We obtain state of the art results for video captioning in challenging Charades dataset obtaining BLEU-4 score of \textbf{34.8} and METEOR score of \textbf{33.6} outperforming previous state-of-the-art of \emph{18.8} and \emph{19.5} respectively. Similarly, on ActivityNet captioning, we obtain excellent results in-terms of ROUGE (20.24) and CIDER (37.58) scores. For action localization, without using any explicit start/end action annotations, our method obtains localization performance of 22.2 mAP outperforming prior fully supervised methods.

\end{abstract}
\section{Introduction}
Human action recognition from videos aims to recognize a \emph{set of predefined human actions} in a given video~\cite{Laptev2005,Wang2013,Simonyan2014,Ji2012}.
To better understand human actions, many related problems have been investigated in the literature, e.g. action detection~\cite{THUMOS14,Yeung2016}, spatial-temporal action localization~\cite{Tian2013}, action segmentation~\cite{Lea2017}, and early action prediction~\cite{Lan2014,Shi2018}.
All these problems involve classifying videos into action categories at some level.
Recently, more challenging human action understanding problems have been proposed such as video captioning~\cite{Yu2016,Mun2019,Venugopalan2015}, text-based temporal activity localization~\cite{Wang2019} and complex activity recognition~\cite{Hussein2019}.
\begin{figure}[t]
\begin{center}
\includegraphics[width=0.99\linewidth]{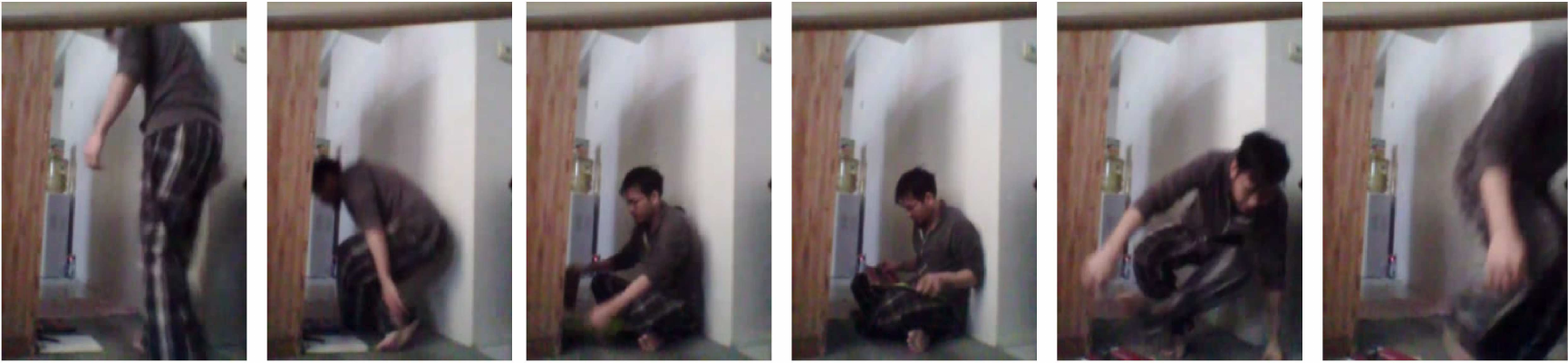}
\end{center}
\caption{If asked to explain the video, one would say ``the man sits down, reads the book and stands up''. It is more human like to describe a video using a sequence of actions.}
\label{fig.video}
\end{figure}

\begin{figure*}[t]
\begin{center}
\includegraphics[width=0.90\linewidth]{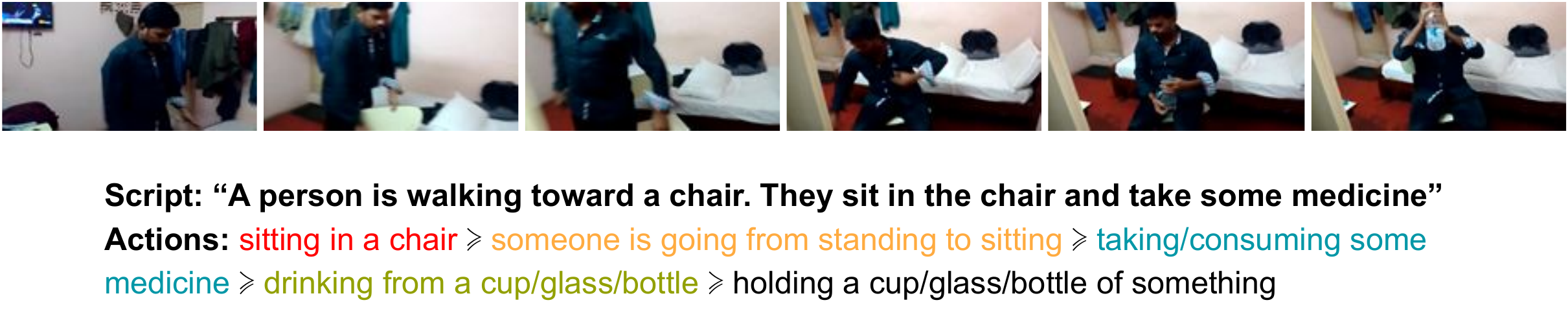}
\end{center}
\caption{Given a complex activity description and the video, \name can also output the sequence of human actions that are required to accomplish the complex activity. This can be used to solve problems such as learning from demonstration.}
\label{fig.application}
\end{figure*}

Most natural videos consist of action sequences and it is also instinctive to describe them using action sequences rather than a set of actions.
If we were asked to explain the activity shown in the video of Figure~\ref{fig.video}, it is likely that we would say ``the man sits down, reads the book and stands up'' in that order which implicitly embodies the temporal order of actions.
This indicates a system that is able to generate an action sequence for a given video is more natural to humans.
Usually, we would not explicitly mention the start and end times of an action as when describing a video as done in activity detection.
Besides, a model's ability to predict sequences of actions from a video has many applications, e.g. learning from demonstration~\cite{Argall2009}. Furthermore, it can be used for video retrieval from temporal action query such as "find videos of player hits the umpire and gets red card" (query $\rightarrow$ \texttt{hit the umpire} $\succ$ \texttt{gets the red card}).
In this paper we analyze the problem where we are given only the video and ground truth action sequence at training time. During the test time, given the video, model should output the correct sequence of actions as humans do. We call this task \emph{\name} and investigate in this paper.

Training a model that is able to output an accurate action sequence without precise temporal annotations is challenging. The model has to learn complex temporal dynamics and relationships between actions of the video before producing the action sequence.
Furthermore, it has to learn implicitly when and where actions do happen and do not only using the input sequence of action labels.
Because one-to-one correspondences between frames and actions are not given, this becomes a difficult learning task.
In a way, model has to align input frame data with semantic action sequence while implicitly learning each human action category.
Ideally, the model should learn complex relationships and inter-dependencies between actions to further improve \name performance.
The most challenging is how to determine the number of actions (that is the length of output action sequence) within the video. 
Too many or less actions in the predicted sequence would significantly hinder the performance.
All of these has to be learned simply using given training data consist of videos and action sequences.

\Name is somewhat related to action detection which involves predicting the start and end of the action along with the confidence of each prediction.
Technically, supervised action detection might be relatively easier task than action-sequence classification during training. 
However, inference for action detection is challenging and therefore, action detectors make use of explicit temporal annotations during training and action proposals for training and inference.
In contrast, we are given only the action sequence without explicit temporal annotations to train our model somewhat similar to weakly supervised action detection~\cite{Paul2018,Nguyen_2018_CVPR,fernando2019weakly,Wang2017}.
However, these methods~\cite{Paul2018,Nguyen_2018_CVPR,fernando2019weakly,Wang2017} neither output sequence of actions nor make use of explicit temporal order of actions during training.
Even though \emph{\name} is more ``human like'' task, it is a challenging one for the machines.

As shown in a recent study, human action boundaries are ambiguous even for humans~\cite{Sigurdsson2017A} and therefore training and evaluation of supervised action detection becomes a challenging task.
In contrast, our task only aims at predicting the sequence of actions and we only penalize for the wrong order of actions ignoring action boundaries explicitly. Therefore, obtaining annotations for our task is somewhat easier, practical and potentially results in consistent and accurate annotations.
A model that is trained to classify action sequences has to learn action boundaries \emph{implicitly}, however, the notion of ``action boundaries'' are not used explicitly during training or testing.
Interestingly, \emph{action-sequence-classification} is useful to solve other downstream problems such as action detection, localization and segmentation.
We illustrate the usefulness of action-sequence classification by solving video-captioning  and action localization problems.
Therefore, we argue that \name is conceptually interesting and practically useful.
\Name can answers the question ``what actions are needed to perform activity X?'' as illustrated Figure~\ref{fig.application}.
Therefore, it is more suitable for some human action understanding tasks~\cite{Argall2009}.

In this paper we propose to tackle a new problem in human action understanding called \emph{human \name}. 
Given a video depicting a complex activity, the objective is to predict the sequence of actions.
In contrast to action classification and detection, this is a sequence-to-sequence learning task.
We propose to solve this problem using machine translation techniques where the input is a video sequence and the output is a sequence of actions.
To summarize, the contributions of this paper are as follows:
\begin{itemize}[leftmargin=*]
\itemsep0em 
 \item We propose a new task in human action understanding called human-action-sequence classification.
 \item We propose a machine translation-based solution to solve this task and investigate two neural translation architectures to solve this challenging problem. 
 \item We evaluate the performance of our solution against several baselines on three datasets, namely the Charades, ActivityNet 1.3 and MPII Cooking datasets and show consistently better results using our model.
 \item We demonstrate usefulness of action sequence classification on two downstream tasks, video captioning and action localization.
 \item We obtain results significantly better than the state-of-the-art on Charades captioning and excellent results on ActivityNet 1.3 caption generation task. 
 \item We obtain better action localization performance outperforming previous supervised methods on Charades dataset. 
\end{itemize}
\section{Related work}
\begin{figure}[t]
\begin{center}
\includegraphics[width=0.99\linewidth]{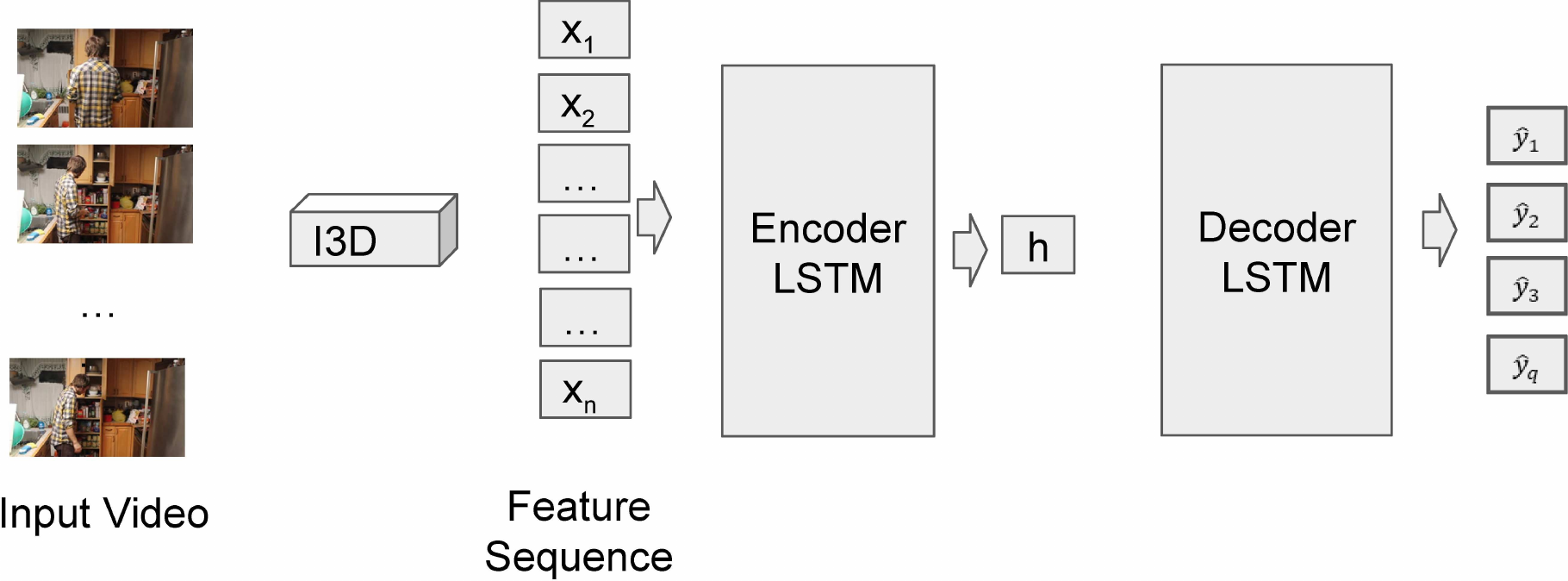}
\end{center}
\caption{Given a complex activity description and the video, action sequence classification outputs the sequence of human actions that are required to accomplish the complex activity.}
\label{fig.overview}
\end{figure}
We propose to tackle action sequence classification problem using LSTM-based machine translation~\cite{Sutskever2014}.
However, typically in machine translation, both the input and output are sequences of words~\cite{Sutskever2014}.
Several CNN-LSTM architectures are proposed to solve various action understanding problems such as action classification, action detection and early action classification.
In~\cite{Du_2015_CVPR} a hierarchical RNN architecture was proposed for skeleton-based action classification.
To improve conventional LSTMs, \cite{Veeriah_2015_ICCV} proposed differential LSTM to make use of spatio-temporal dynamics for video action classification. 
LRCN~\cite{donahue2015long} is the one of the first methods to use LSTMs for action classification and the video captioning. Their video captioning solution is a sequence-to-sequence one, however they do not use encoder-decoder architecture as we do; besides video captioning is different from action sequence classification task.

The work known as ``Watch-n-Patch'' \cite{Wu_2015_CVPR} is somewhat related to us as it attempts to understand a sequence of actions in an unsupervised manner to predict the missing action.
Similarly, \cite{Xu_2015_ICCV} propose an activity auto-completion (AAC) model for human activity prediction by formulating activity prediction as a query auto-completion (QAC) problem in information retrieval using learning to rank. 
However, they do not investigate the problem of generating action-sequences for a given video. 
Instead, they aim to predict the next action that is going to happen in the future. 
Context-aware \cite{Hasan_2015_ICCV} action recognition is also related to us as it attempts to explore contextual information for action recognition. 
Our method make use of entire video to predict the action sequence by aligning the input video sequence with the output action-sequence using an attentional mechanism. 
Our problem and the solution is a sequence-to-sequence one while the method in \cite{Hasan_2015_ICCV} study the problem of action classification which predicts a set of actions within a video.

A bi-directional RNN is used for action detection in~\cite{Singh_2016_CVPR}. Like most other methods that uses LSTMs/RNNs for action understanding tasks, this method also takes the video sequence as input and produces a sequence of action prediction for each frame or segment. RNN model is trained with one-to-one input-output sequence correspondences. If the input video sequence has $n$ number of elements, usually most action recognition methods that uses RNNs would output $n$ number of action predictions and then aggregate that information to make the final action classification prediction~\cite{Singh_2016_CVPR,donahue2015long}. However, for us, the input and the output sequence sizes and dimensions are different.
The most similar to use is the work of~\cite{Liu2019}. They also make use of encoder-decoder architecture for event-detection in videos. However, they apply mean pooling over the decoder to obtain event prediction and therefore not generating a sequence of events/actions for a given video. Therefore, our work is different from them. 
Similarly, LSTM-based future actions of a video sequence prediction is presented in~\cite{gammulle2019forecasting}. Despite the title, this paper does not output a sequence of actions, but outputs an action for each future frame. However, interestingly, their encoding process is somewhat similar to us.
Our objective is to demonstrate the value of a model that is able to output an action sequence at test time for a given video. We demonstrate that it is useful to solve many downstream video understanding tasks such as action detection, segmentation, localization and video captioning. Furthermore, it can be useful for practical applications such as learning from demonstration. Therefore, our work conceptually and technically differ from these work~\cite{Liu2019,gammulle2019forecasting}.

Our approach to action understanding differs from main stream action detection~\cite{Yeung_2016_CVPR,Singh_2016_CVPR} and action segmentation~\cite{shi2008discriminative} due to the nature of supervision used. The output of these methods can be further processed to align with the action-sequence, e.g. using clustering. However, these methods use precise temporal annotations during training and therefore different from our model and the task. Perhaps weakly supervised action segmentation is the closest to our problem~\cite{Chang_2019_CVPR}. However, weakly supervised action segmentation is more challenging than our problem as it needs to infer temporal boundaries only using action-sequence. Similarly, in weakly supervised action detection, non of the methods can generate a sequence of actions without further processing and besides they do not make use of chronological order of actions during training~\cite{Paul2018,Nguyen_2018_CVPR,fernando2019weakly,Wang2017}.

\section{Action sequence classification}
\subsection{Problem}
Given a RGB video sequence $X=\left<x_1, x_2, \cdots x_n \right>$ and the corresponding sequence of human actions $Y=\left< y_1, y_2, \cdots y_p \right>$, we learn a model that generates the action sequence $Y$ from the video sequence $X$. 
Here $x_i$ is a RGB frame and $y_j$ is a categorical human action. 
$\mathcal{Y}$ is the set of human actions and each action $y_j \in \mathcal{Y}$.
The total number of human actions is fixed, i.e. $|\mathcal{Y}|=C$.
Then the model objective is to learn a set of parameter $\Theta$ such that it can predict the action sequence as follows:
\begin{equation}
 \left< y_1, y_2, \cdots y_p \right> = \Phi(\left<x_1, x_2, \cdots x_n \right>,\Theta)
\end{equation}
where both input sequence $X$ and the output sequence $Y$ of arbitrary length.
This is a sequence to sequence machine translation task~\cite{Sutskever2014} where the input sequence consists of three dimensional tensors (RGB frames) and the output sequence consists of categorical symbols (action classes).

\subsection{High level idea}
To solve this problem, first we extract a sequence of features from each input RGB video sequence.
Recently, there has been some significant work in action recognition including methods such as inflated 3D convolutions (I3D)~\cite{Carreira2017}, temporal relation networks~\cite{Zhou2018}, and temporal segment networks~\cite{Wang2016}. 
We use I3D features as the video clip representation and obtain a sequence of I3D features from the input video $X$ due to its good temporal footprint and good performance.
Let us denote the sequence of visual features obtained for each video by $\mathbf{X} = \left< \mathbf{x_1}, \mathbf{x_2}, \cdots \mathbf{x_T} \right>$.
Then we use Long Short Term Memory (LSTM)~\cite{Hochreiter1997} networks to encode I3D visual feature sequence to obtain a single vector representation (e.g. hidden state of LSTM) after processing the entire video.
Afterwards, the LSTM decoder network  takes the hidden state of encoder LSTM as the initial state to generate the output sequence $Y$.
The high level idea of our method is shown in Figure~\ref{fig.overview}.
Next we explain two state-of-the art machine translation networks and adapt them to solve our video-to-action-sequence problem.

\subsection{LSTM-Encoder-Decoder: Sequence to sequence model}
\label{sec.basic}
\begin{figure}[t]
\begin{center}
\includegraphics[width=0.99\linewidth]{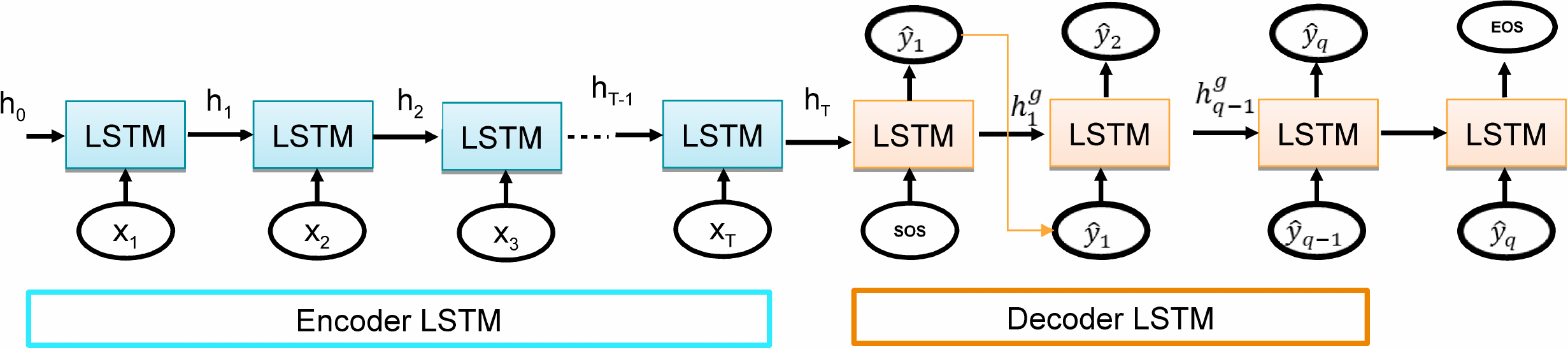}
\end{center}
\caption{A visual illustration of our LSTM Encoder-Decoder architecture for video feature sequence to action sequence translation.}
\label{fig.LED}
\end{figure}

In this section we explain our basic sequence to sequence model coined \emph{LSTM-Encoder-Decoder} which takes a sequence of image features $\mathbf{X} = \left< \mathbf{x_1}, \mathbf{x_2}, \cdots \mathbf{x_T} \right>$ as input and return a sequence of actions $Y=\left< y_1, y_2, \cdots y_p \right>$.
Indeed, our model is adapted from state-of-the-art machine translation architecture proposed in~\cite{Sutskever2014}.
This LSTM-Encoder-Decoder architecture is shown in Figure~\ref{fig.LED}.
%
%
%
Let us denote the encoder LSTM by $f()$ defined as follows:
\begin{equation}
\mathbf{c_t},\mathbf{h_t} = f(\mathbf{x_t},\mathbf{h_{t-1},\mathbf{c_{t-1}}}) 
\label{eq.encoder}
\end{equation}
where $\mathbf{c_t}$ is the cell state and the $\mathbf{h_t}$ is the hidden state of the encoder LSTM. 
For a given input feature sequence $\mathbf{X}$, the final cell and hidden state of the encoder LSTM is denoted by $\mathbf{c_T}$ and $\mathbf{h_T}$ respectively.
The decoder LSTM has similar structure to the encoder LSTM, however it takes the previously predicted action class (symbol) as input $\hat{y}_{q-1}$ and returns the next action symbol $\hat{y}_q$.
The initial token $<$SOS$>$ is the first input to the decoder.
The decoder LSTM has two distinct linear mappings.
The first one takes one-hot vector representation of the action symbol $y_q$ (including $<$SOS$>$) and returns a vector representation of that by learning an embedding matrix. Let us denote this mapping by 
\begin{equation}
\mathbf{y_q} = \phi(y_q;W_{emb})
\label{eq.emb}
\end{equation}
where $W_{emb}$ is the embedding parameter.
Therefore, this operation (equation~\ref{eq.emb}) maps the input symbol $y_q$ to a $D$ dimensional vector.
Therefore the actual input sequence to the decoder is $\left< \mathbf{\hat{y}_1}, \mathbf{\hat{y}_2}, \cdots, \mathbf{\hat{y}_p} \right>$ during training. Let us denote the decoder LSTM by $g()$ defined as follows:
\begin{equation}
\mathbf{s_q}, \mathbf{h^g_q} = g(\mathbf{\hat{y}_{q-1}}, \mathbf{s_{q-1}},\mathbf{h^{g}_{q-1}}) 
\label{eq.decoder}
\end{equation}
where $\mathbf{s_q}$ is the context state of the decoder and $\mathbf{h^g_q}$ is the hidden state.
Decoder outputs a sequence of actions using the second linear mapping that maps hidden state to obtain the next symbol as follows: 
\begin{equation}
\hat{y}_q = \mathtt{argmax} W \mathbf{h^g_q}
\end{equation}
where $W$ is learned.
The final hidden state (and cell state) of the encoder is used to initialize the initial hidden state (and cell state) of the decoder LSTM i.e. $\mathbf{h^g_0} = \mathbf{h_T}$ and $\mathbf{s_0} = \mathbf{c_T}$.

Both encoder and the decoder is trained by minimizing the cross-entropy loss between $(y_q, \hat{y}_q)$.
We train our model with end-of-sequence token $<$EOS$>$ to determine the length of the target sequence.
During inference, we stop generating as soon as we generate the $<$EOS$>$ token.
This means, the decoder has two additional symbols, the initial token $<$SOS$>$ and the end token $<$EOS$>$.
We also use teacher forcing strategy during training where the ground truth symbol vector $\mathbf{y_{q-1}}$ is fed to the decoder in equation \ref{eq.decoder} instead of predicted output $\mathbf{\hat{y}_{q-1}}$. 
We use this strategy 50\% of the time (at random) during training.

\subsection{GRU-AA: GRU alignment and attention}
\label{sec.gru}
In section~\ref{sec.basic}, the decoder relies only on the final encoding of the encoder LSTM (i.e. $\mathbf{h}_T$ and $\mathbf{c}_T$) to produce the output sequence.
Perhaps, it is more intuitive to find the most relevant set of output states of the encoder that generates the output symbol $y_q$.
One way to do this is to use attention mechanism over the encoder LSTM outputs as done in~\cite{bahdanau2014neural}.
Furthermore, instead of using LSTM, we use a GRU which simplifies the encoder $f()$ as follows: 
\begin{equation}
\mathbf{h_t} = f(\mathbf{x_t},\mathbf{h_{t-1}}) .
\label{eq.encoder_gru}
\end{equation}
We also use a GRU for the decoder.
For a given input feature vector sequence $\mathbf{X}=\left< \mathbf{x_1}, \mathbf{x_2}, \cdots, \mathbf{x_T} \right>$, the encoder GRU $f()$ produces the sequence of hidden states $\mathbf{H}=\left< \mathbf{h_1}, \mathbf{h_2}, \cdots, \mathbf{h_T} \right>$.
To produce the decoder output $\mathbf{\hat{y}_q}$,  GRU-AA model learns attention weights over the encoder hidden state sequence $\mathbf{H}$.
The weight ($\beta_i$) assigned to the encoder hidden state $\mathbf{h_i}$ for generating symbol $\hat{y}_q$ is obtained by the following:
\begin{equation}
\beta_i =  tanh([\mathbf{h_i}; \mathbf{h^g_q}]^T \times W_{att}) \times V
\label{eq.att}
\end{equation}
where $V{}$ is a learnable parameter of size $\mathcal{R}^{1\times D}$ and $W_{att}$ is the learnable attention matrix of size $\mathcal{R}^{2D\times D}$. 
The notation $[\mathbf{h_i}; \mathbf{h^g_q}]$ is used for column vector concatenation.
Thereafter, to obtain the attention weight for encoder hidden state $\mathbf{h_i}$ for generating action symbol $y_q$, we use softmax function over all weights $\{ \beta_1, \beta_2, \cdots, \beta_T \}$ as follows:
\begin{equation}
\alpha_{i,q} = att(\mathbf{h_i},\mathbf{h^g_q}) =  \frac{exp(\beta_i)}{\sum_{j=1}^{T} exp(\beta_j)}.
\label{eq.att2}
\end{equation}
As only a handful of hidden states in sequence $\mathbf{H}$ contributes to generate the output symbol $y_q$, it makes sense to use attention over $\mathbf{H}$ when generating the next symbol using the decoder $g()$.
To do that, we propose to compute a context vector which is a weighted sum of encoder hidden states where the weight is given by equation~\ref{eq.att2}. 
For generating $q^{th}$ action symbol, then the context vector is obtained by equation~\ref{eq.con}.
\begin{equation}
 \mathbf{c^g_{q-1}} = \sum_{j=1}^{T} \alpha_{i,{q-1}} \mathbf{h_{j}}.
 \label{eq.con}
\end{equation}
After that, we modify the decoder GRU to take this context vector along with the action sequence vector $\mathbf{\hat{y}_{q-1}}$ as follows:
\begin{equation}
\mathbf{h^g_q} = g([\mathbf{\hat{y}_{q-1}};\mathbf{c^g_{q-1}}], \mathbf{h^{g}_{q-1}}) 
\label{eq.decoder_gru}
\end{equation} 
where $[\mathbf{\hat{y}_{q-1}};\mathbf{c^g_{q-1}}]$ is the vector concatenation.
To obtain the next action symbol, we use the following linear mapping ($U$) over three concatenated vectors, i.e., the hidden state of the decoder, the attention weighted context vector and the previous action vector representation.
\begin{equation}
\hat{y}_q = \mathtt{argmax} U [\mathbf{h^g_q}; \mathbf{c^g_{q-1}}; \mathbf{\hat{y}_{q-1}}]
\label{eq.yhat}
\end{equation}

Overall, our model has the four parameters to learn apart from the GRU/LSTM parameters.
They are the embedding parameter $W_{emb}$, the attention matrix $W_{att}$, the attention parameter $V$, and the output linear mapping parameter $U$.
All these parameters are learned with the cross-entropy loss.
We also use teacher forcing strategy as before to train GRU-AA model to translate video sequences to action sequences.

\subsection{Video captioning}
\label{sec.vid}
\begin{figure*}[t]
\begin{center}
\includegraphics[width=0.85\linewidth]{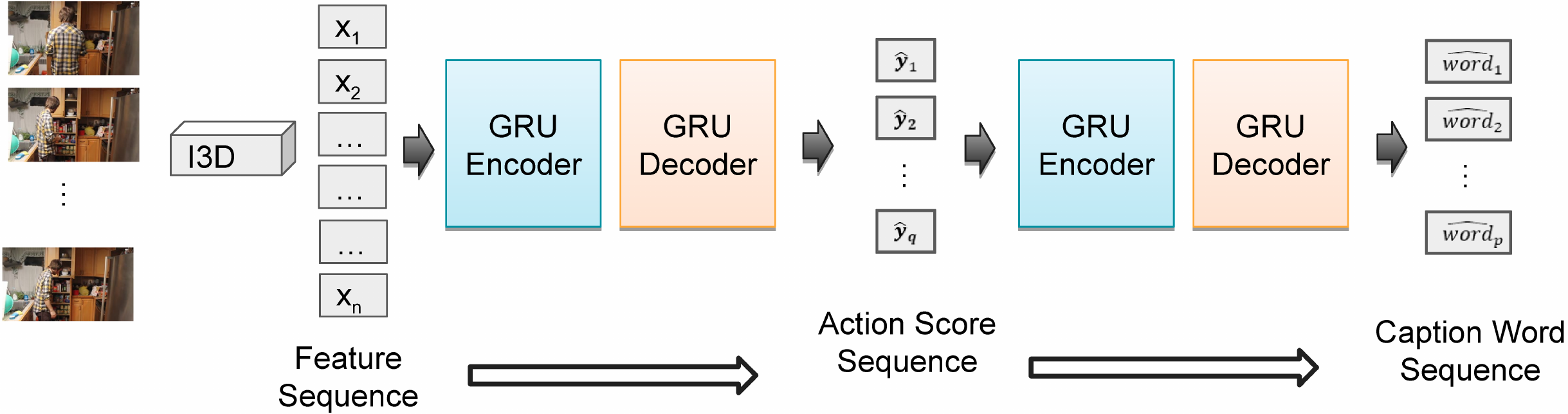}
\end{center}
\caption{A visual illustration of our captioning architecture that uses two GRU-AA models. The first one takes visual feature sequence as input and outputs a sequence of action predictions using the model presented in Section~\ref{sec.gru}. Then the second GRU-AA Seq.-to-Seq. model takes this sequence of action predictions as input and outputs the sequence of words.}
\label{fig.caption.archi}
\end{figure*}
Our method can be used to solve other downstream tasks such as action detection and action segmentation.
Here we use our method for solving video captioning which we will evaluate in the experiments section.
Let us assume that we are given the sequence of video features $\mathbf{X}$ and the corresponding captions.
Captions are a sequences of words.
Using Glove~\cite{pennington2014glove} vector representation, we transform captions (word sequences) into a sequence of Glove vectors.
Then using our GRU-AA encoder-decoder model, we translate the video feature sequence into an action-prediction-sequence. 
Let us denote this encoder-decoder by $H_{X2Y}$.
Instead of generating actions symbols by taking the argmax as in equation~\ref{eq.yhat}, for $H_{X2Y}$, we directly take the output predictions (score vectors) and generate the sequence of action predictions. 

Then using another GRU-AA encoder-decoder model $H_{Y2W}$, we translate the sequence of action predictions into Glove word-sequence.
Therefore, we have two Sequence-to-Sequence models (namely $H_{X2Y}$ and $H_{Y2W}$) in our video captioning architecture as illustrated in Figure~\ref{fig.caption.archi}.
During training, we use captions, action-sequences and video features. 
During testing we generate captions only using video features.
First we train action sequence generation model as in Section~\ref{sec.gru} and then use that model to initialize  $H_{X2Y}$. 
Then we jointly train $H_{X2Y}$ and $H_{Y2W}$ to generate captions.
\section{Experiments}

\subsection{Dataset}
To evaluate the video-to-action-sequence classification, we need a dataset that consists of multiple actions within a video.
Therefore, we use three datasets, the Charades dataset \cite{sigurdsson2016hollywood}, MPII Cooking dataset~\cite{Rohrbach2012} and the ActivityNet 1.3 dataset \cite{FabianCabaHeilbron2015}. 

In Charades dataset, each video has a list of action labels accompanied by corresponding timestamps. The timestamps are not used in the experiments explicitly. Timestamps were used to generate the sequence of actions for each video. Charades dataset has 7,985 video for training and 1,863 videos for testing.  The dataset is collected in 15 types of indoor scenes, involves interactions with 46 object classes and has a vocabulary of 30 verbs leading to 157 action classes~\cite{sigurdsson2016hollywood}. On average there are  6.8 actions per video which is much higher than most other datasets having more than 1000 videos.

MPII-Cooking Dataset~\cite{Rohrbach2012} has 65 fine grained actions, 44 long videos with a the total length of more than 8 hours. Twelve participants interact with different tools, ingredients and containers to make a cooking recipe.  We use the standard evaluation splits where total of five subjects are permanently used in the training set.  Rest six of seven subjects are added to the training set and all models are tested on a single subject and repeat seven times in a seven fold cross-validation manner. In this dataset, there are 46 actions per video on average, much larger than other datasets.

ActivityNet 1.3 dataset aims at covering a wide range of complex human activities that are of interest to people in their daily living. This dataset consists of 200 action classes and on average 2 actions per video~\cite{FabianCabaHeilbron2015}.

\subsection{Evaluation matrix}
As this is a sequence evaluation task, we use Bilingual Evaluation Understudy (BLEU) score~\cite{Papineni02bleu:a}.
Specifically, we use BLEU-1 and BLEU-2 scores as some datasets contains at most two actions per video.
We also report sequence-item classification accuracy which counts how many times the predicted sequence elements match the ground truth in the exact position. For example, if the ground truth is $\left< y_a,y_b,y_a \right>$ and the predicted is $\left< y_a,y_b,y_a,y_c \right>$, as there are three elements that exactly matches the accuracy would be 3/4*100\%. 
If the predicted is $\left< y_a,y_b \right>$ then accuracy is 2/3*100\% and if the output is $\left< y_b, y_a, y_b, y_a \right>$ accuracy is 0\% as none of the elements match with the ground truth.

\subsection{Implementation Details}
We use I3D network \cite{Carreira2017} which is pre-trained on Kinetics dataset~\cite{Kay2017} for action classification.
The hidden size used for models with a single input is 512. The embedding size used for the action tokens is 512 (i.e. output size of $W_{emb}$). We train our models with $<$SOS$>$, $<$EOS$>$ and padding symbols.
Action sequence classification models are trained with batch size of 32 for 10 epochs using a learning rate of $1e^{-3}$ with early stopping.

\subsection{Baseline comparisons}
\label{sec.base}
We compare our LSTM:Encoder-Decoder (\textbf{LSTM-ED})  (Section~\ref{sec.basic}) and \textbf{GRU-AA} (Section~\ref{sec.gru}) with three baselines. First, we evaluate against the random performance. Second, we use the mean pooled I3D features as input and use a stacked LSTM model (\textbf{LSTM-Mean}) (structure is similar to our model) to output the sequence of action symbols. Third, we train a LSTM model (\textbf{LSTM-SS}) that also use sequence of I3D features as the input and then output the action sequence as the output. 
Results are reported in Table~\ref{tbl.main}.
\begin{table}[t]
\begin{center}
\begin{small}
\begin{tabular}{|c|c|c|c|}
\hline
Model & Class. Acc. (\%) & BLEU-1 & BLEU-2 \\\hline 
\multicolumn{4}{|c|}{Charades dataset.} \\ \hline
Random 		& 0.35 		& 2.02 		& 0.02 \\ 
LSTM-Mean 	& 4.54 		& 9.64 		& 1.03 \\
LSTM-SS 	& \B{5.13} 	& 10.69 	& 1.12 \\
LSTM-ED  	& 3.18 		& 10.63 	& \B{1.62} \\ 
GRU-AA  	& 4.64 		& \B{14.83} 	& \B{1.62} \\  \hline
\multicolumn{4}{|c|}{MPII Cooking.} \\ \hline
LSTM-Mean 	& 11.17 	& 15.47 	& 8.75 \\ 
LSTM-SS 	& 14.06		& 19.92		& 10.65 \\
GRU-AA  	& \B{14.86} 	& \B{25.74} 	& \B{14.34} \\ \hline
\multicolumn{4}{|c|}{ActivityNet 1.3} \\ \hline
LSTM-Mean 	& 34.43		& 38.09		& 2.95 \\ 
LSTM-SS 	& 10.70		& 12.31		& 0.50 \\
GRU-AA  	& \B{45.0}	& \B{51.53}	& \B{3.64} \\ \hline
\end{tabular}
\end{small}
\end{center}
\caption{Comparison of results for action sequence classification task using I3D features on three action recognition datasets.}
\label{tbl.main}
\end{table}

First, from these results we see that all models performs better than random performance.
On average the results on Charade dataset is the lowest indicating the difficulty of the task on this challenging dataset.
All models obtain the best results on ActivityNet dataset.
Interestingly, LSTM-ED and GRU-AA method outperform other baselines except on Charades dataset where LSTM-SS baseline obtains sequence classification accuracy of 5.13. On all other datasets the best performer is GRU-AA which indicates that our design choice is the correct one for this challenging task.
Somewhat unexpectedly, LSTM-Mean model also obtains good results perhaps suggesting the power of I3D features.
Interestingly, our method (GRU-AA) obtains significantly better results than other baselines on ActivityNet dataset while obtaining a BLEU-1 score of 51.33. However the BLUE-2 score is still relatively lower than one would like yet better than other baselines.
Nevertheless, still ours is able to obtain a BLEU-2 score of 14.34 on MPII Cooking dataset where there are more actions in each video.
Results from MPII Cooking dataset suggest that ours is the best method by a large margin in terms of BLEU scores.
From all these results we conclude that the best performing method is our GRU-AA model.

\subsection{Downstream Task 1: Video captioning}
\begin{figure}[t]
\begin{center}
\includegraphics[width=0.75\linewidth]{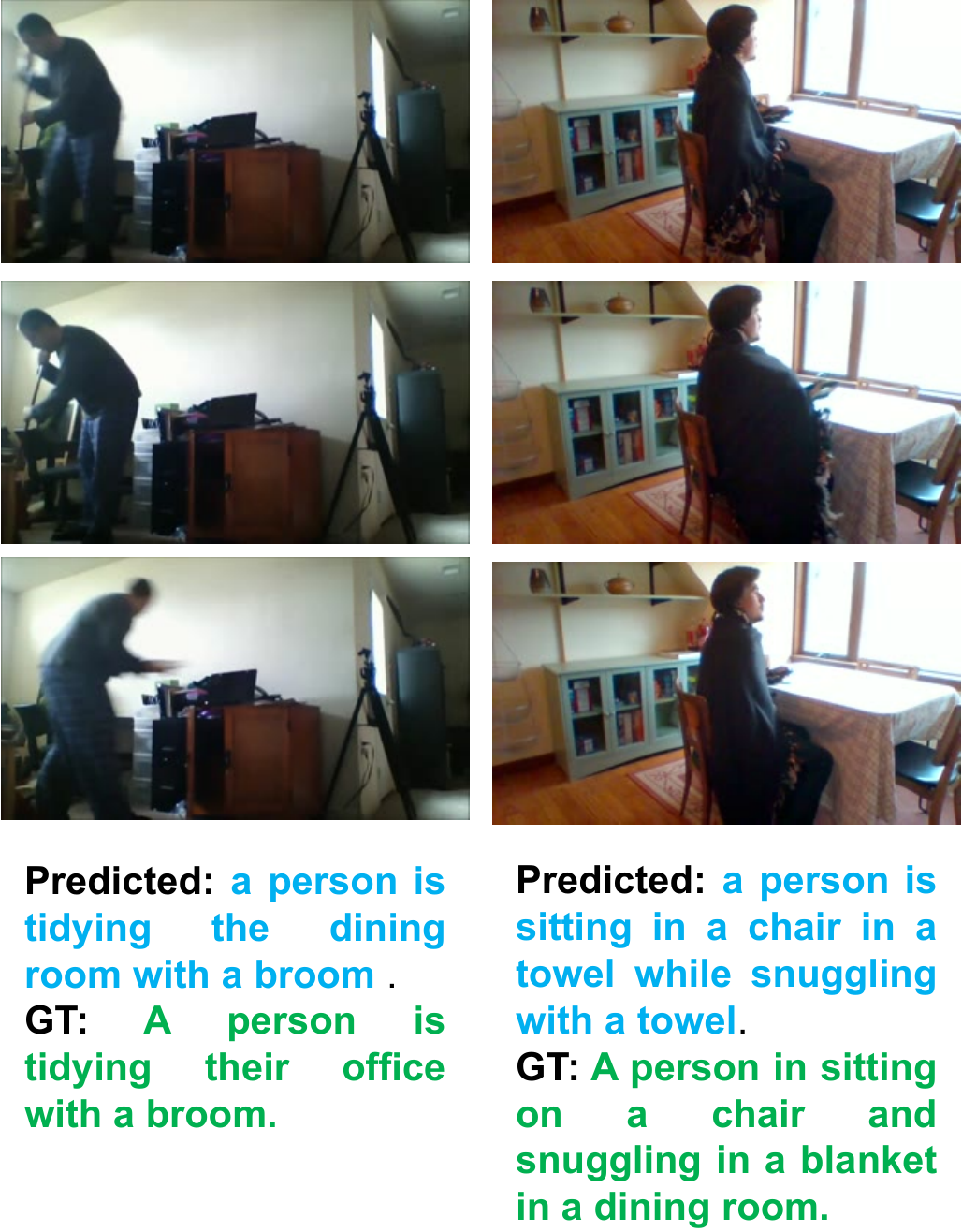}
\end{center}
\caption{Generated captions using our model. Ground truth (GT) captions are shown in green and the predicted ones are in blue.}
\label{fig.cap.res}
\end{figure}
Now we evaluate our video captioning method presented in Section~\ref{sec.vid} using Charades and ActivityNet1.3 dense captioning datasets~\cite{Krishna2017}.
We use the I3D feature sequence to first generate the sequence of actions (score sequence) and then use another sequence-to-sequence model (GRU-AA) to translate the sequence of action-predictions into captions.
Therefore, our \B{Stage-GRU-AA} model consists of two stages of GRU-AA models as explained in section~\ref{sec.vid}.

We use standard captioning protocol and data as in~\cite{Zhao2018,Wu_2018_CVPR,Fakoor2016,li2017mam} for Charades.
Recently, authors of~\cite{Wang_2018_CVPR} introduced a new set of captions for Charades dataset called ``Charades captions'' which we also use.
The most other recent methods~\cite{Zhao2018,Wu_2018_CVPR} use the original Charades captions and  we also used the exact data as used in~\cite{Zhao2018,Wu_2018_CVPR,Fakoor2016,li2017mam}.
The dataset is split into three parts, including 7985 for training, of which 30\% is used for validation, and 1863 for testing. 
We use the protocol of~\cite{Krishna2017} for ActivityNet captioning.

First we compare all baseline models using both Charades and ActivityNet 1.3 dense captioning in Table~\ref{tbl.caption.baseline}.
The first LSTM baseline takes the mean I3D features as input and generates captions for the video using a stacked LSTM (\B{LSTM-Mean}).
The second LSTM baseline model takes the sequence of I3D features as input and generates the captions using a stacked LSTM denoted by (\B{LSTM-SS}).
The third baseline is an LSTM:Encoder-Decoder (\B{LSTM-ED}) model presented in section~\ref{sec.basic}. Here the model takes the sequence of I3D features as input and generates the output captions. 
The fourth baseline is a  \B{GRU-AA} model presented in section~\ref{sec.gru} that takes the sequence of I3D features and outputs the captions. We use the same GRU encoder-decoder with attention model as described in section~\ref{sec.gru}.
\begin{table}[t]
\begin{center}
\begin{small}
\begin{tabular}{|l|c|c|c|c|c|}\hline
Method & METEOR & ROUGE-L  \\ \hline
\multicolumn{3}{|c|}{Charade captions dataset.}\\ \hline
LSTM-Mean 	& 22.76 &	29.49 \\
LSTM-SS   	& 22.45 &	28.70 \\
LSTM-ED   	& 19.08 &	29.22 \\
GRU-AA 	  	& 22.34 &	32.61\\
Stage-GRU-AA	& \B{33.60} &	\B{38.15}\\  \hline
\multicolumn{3}{|c|}{ActivityNet1.3 captions dataset.}\\ \hline
LSTM-Mean 	& 6.12	& 	17.94 \\
LSTM-SS   	& 5.38	&	16.34 \\
Stage-GRU-AA	& \B{8.24}& 	\B{20.23}\\  \hline
\end{tabular}
\end{small}
\end{center}
\caption{Charades video captioning results using our method presented in Section~\ref{sec.vid}.}
\label{tbl.caption.baseline}
\end{table}

From Table~\ref{tbl.caption.baseline}, we see that LSTM-Mean model performs quite reasonably despite it's simplicity.
Somewhat surprisingly, the LSTM-SS model shows decrease in performance in both datasets. LSTM-ED model that uses LSTM-based encoder-decoder for video caption generation obtains lower results in METEOR but somewhat moderate results for ROUGE-L. GRU-AA model that takes the visual feature sequence and translates to captions seems to work relatively better than all previous methods indicating that attention is important for this task.
Our Stage-GRU-AA model that first translate the videos to action sequence and then action sequence to captions obtains the best results in both datasets. The improvement is quite significant. This shows the importance of learning this intermediate high-level semantic representation of temporal information in the form of action sequence for downstream task of video captioning.

Next we compare our model performance with state-of-the-art methods and report results in Table \ref{tbl.caption} for challenging Charades dataset.
\begin{table}[t]
\begin{center}
\begin{small}
\begin{tabular}{|l|c|c|c|c|c|}\hline
Method & B1 & B2 & B3 & B4 & M \\ \hline
\multicolumn{6}{|c|}{Charade dataset results}\\ \hline
S2VT~\cite{Venugopalan_2015_ICCV} ICCV15 & 49.0 & 30.0 & 18.0 & 11.0 & 16.0  \\
SA~\cite{Yao_2015_ICCV} ICCV15 & 40.3 & 24.7 & 15.5 & 10.8 & 14.3 \\
MAAM~\cite{Fakoor2016} 2016 & 50.0 & 31.1 & 18.8 & 11.5 & 17.6\\
MAM~\cite{li2017mam} IJCAI17 & \textbf{53.0} & 31.7 & 21.3 & 13.3 & 19.1  \\
TSL~\cite{Wu_2018_CVPR} CVPR18 & -- & --& -- & 13.5 & 17.8  \\
VCTF~\cite{Zhao2018} ICJAI18 & 50.7 & 31.3 & 19.7 & 13.3 & 19.0  \\ 
Our & 50.4& \textbf{45.6} & \textbf{41.0} & \textbf{33.5} & \textbf{23.4} \\ \hline
\multicolumn{6}{|c|}{Charade captions* dataset results}\\ \hline
HRL~\cite{Wang_2018_CVPR} CVPR18 & \textbf{64.4} & 44.3 & 29.4 & 18.8 & 19.5  \\
Our & 53.0& \textbf{48.2} & \textbf{42.9}& \textbf{34.8} & \textbf{33.6} \\
\hline
\end{tabular}
\end{small}
\end{center}
\caption{Charades video captioning results using our method presented in Section~\ref{sec.vid}. B1-B4 stands for BLEU-1 to BLEU-4 scores and M stands for METEOR.}
\label{tbl.caption}
\end{table}
We can see immediately the impact of our method. We outperform all other methods in most measures (BLEU-2 to BLEU-4 and METEOR).
Our results are quite significant for BLEU-4 and METEOR where we outperform recent methods such as TSL~\cite{Wu_2018_CVPR} and HRL~\cite{Wang_2018_CVPR} by almost \textbf{20.0} BLEU-4 points. 
Similarly, our METEOR score is \textbf{4.3} points better than the best performing method. 
We significantly outperform HRL~\cite{Wang_2018_CVPR} in METEOR score.

We believe the reason for our excellent results are three fold, i. action sequence prediction model captures the temporal evolution of human actions and keep relevant information to summarize the activity in the video, ii. our GRU-AA machine translation model is capable of effectively aligns and translate input sequence to output, and iii. the decision to use action sequence predictions scores instead of the action sequence is also crucial as score distribution contains more information and robust.

These results indicates the effectiveness of action sequence prediction for solving downstream task/problems such as video captioning.
We also visualize some of the generated captions of our model in Figure~\ref{fig.cap.res}.
Our method is able to accurately generate captions containing the correct actions.
Interestingly, it is not able to correctly identity the context (dining room vs office and for the second one context is missing).
This is not surprising as our method is trained to predict actions accurately but not the context.
In future, we aim to investigate how to include more contextual information to improve video captioning task.
We also report dense captioning results using ActivityNet 1.3 dataset in Table~\ref{tbl.caption.act3} using ground truth (GT) proposal and compare with other methods that used GT proposals.
We obtain good results in-terms of CIDER and ROUGE-L. However, the METEOR score is relatively low for us.
This might be due to the fact that there are only ~2 actions per video in ActivityNet whereas Charades has more actions and our method can better exploit those to generate better captions.
Here our intension is to show that action sequence classification is useful for other downstream tasks.
We hope to improve ActivityNet captioning which we leave for future work where there are only few actions per video.
\begin{table}[t]
\begin{center}
\begin{small}
\begin{tabular}{|l|c|c|c|c|}\hline
Method & M & C & R  \\ \hline
Dense\cite{Krishna2017} 	& 9.46 		& 24.56 	& -- \\
Mask~\cite{Zhou2018a} 		& \B{11.10} 	& -- 		& -- \\
JEDDi~\cite{Xu2019} 		& 8.58 		& 19.88 	& 19.63 \\
DVC~\cite{Li2018} 		& 10.33 	& 25.24 	& -- \\
Our 				& 8.25 		&\B{37.58} 	& \B{20.24} \\ \hline
\end{tabular}
\end{small}
\end{center}
\caption{ActivityNet dense captioning results using our method presented in Section~\ref{sec.vid}. M stands for METEOR, R for ROUGE-L and C for CIDER.}
\label{tbl.caption.act3}
\end{table}
\subsection{Downstream Task 2: Action localization}
Our second downstream task is action localization in Charades dataset. We use GRA-AA model trained in section~\ref{sec.base} solve this. Following action localization protocol in~\cite{Piergiovanni2018,fernando2019weakly,Sigurdsson2017B}, we classify 25 equal distant frames and obtain the localization scores and mAP. As we do not use precise temporal annotations (as done in~\cite{Piergiovanni2018,Sigurdsson2017B}) to train, and therefore, it is a \emph{moderately supervised} method. We report results in Table~\ref{tab-soa}.
\begin{table}[t]
\centering
\scriptsize{
\begin{tabular}{|l|l|l|l|} \hline
Supervision & Method & Features & mAP \\ \hline
\multirow{6}{*}{Supervised} &Temporal Fields~\cite{Sigurdsson2017B} &   VGG16 (RGB+OF)	 	& 12.8\\ 
& Two Stream++~\cite{Simonyan2014}         & VGG16 (RGB+OF)  	& 10.9\\ 
&Super-Events~\cite{Piergiovanni2018} & I3D (RGB)     	& 18.6 \\ 
&Super-Events~\cite{Piergiovanni2018} & I3D  (RGB+OF)   	& 19.4 \\ 
&LSTM & I3D (RGB+OF)					& 18.1 \\ 
&TGM \cite{Piergiovanni2019} &  I3D (RGB+OF)				& 21.5 \\ \hline
Weakly&WSGN~\cite{fernando2019weakly}& I3D (RGB)			& 18.3\\ \hline
\multirow{2}{*}{Moderately}&Our&  I3D (RGB)						& 20.1 \\
&Our & I3D (RGB+OF)						& \B{22.2} \\\hline
\end{tabular}
}
\vspace{0.1cm}
\caption{Compare to state-of-the for action localization on Charades. Methods use only RGB and Optical flow (OF).}
\label{tab-soa}
\end{table}
Compared state-of-the-art methods such as Super-Events~\cite{Piergiovanni2018} and TGM \cite{Piergiovanni2019}, we obtain quite reasonable results without using \emph{precise temporal annotations} during training. It is also better than recent weakly supervised WSGN~\cite{fernando2019weakly}.
Our GRU-AA outperforms new TGM \cite{Piergiovanni2019} \footnote{combination of TGM and Super-Events obtain 22.3~\cite{Piergiovanni2019}.} method by 0.7 map.
This is because our method explicitly model correlations between actions and better align action sequence with the feature sequence using an attensional mechanism.
These results suggest the power of action sequence classification task for solving downstream tasks such as action localization and video captioning.

\section{Discussion and Conclusion}
In this paper we presented a new task called video action sequence classification to output a sequence of actions instead of set of action as done in action recognition literature.
We formulate two solutions to this problem using neural machine translation.
We obtain interesting and encouraging results on three difficult action recognition datasets, the MPII Cooking, Charades and ActivityNet datasets.

Secondly, we extend action sequence classification for solving other downstream tasks, specifically, for video captioning and action localization.
We obtained significant improvements over prior state-of-the art results in video captioning and action localization on Charades dataset. 
Similarly, we obtain significant results on ActivityNet captioning for CIDEr and ROUGE-L.
In the future, in an extended paper we demonstrate on other downstream tasks to convince the community that \emph{"action sequence classification"} is worthy of investigating. Finally, we will release the codes and model for the community to make use of them.
\\\small{\textbf{Acknowledgment:}This research is supported by the National Research Foundation Singapore under its AI Singapore Programme (Award Number: AISG-RP-2019-010).}
{\small
\bibliographystyle{ieee}
\bibliography{egbib}
}
\end{document}